\documentclass[journal,twoside,web]{ieeecolor}
\usepackage{generic}
\usepackage{cite}
\usepackage{bm}
\usepackage{multirow}
\usepackage{amsmath,amssymb,amsfonts}
\usepackage{algorithmic}
\usepackage{graphicx,lipsum}
\usepackage{textcomp}
\usepackage{hyperref}
\usepackage{booktabs}
\usepackage{multirow}
\usepackage[table]{xcolor}
\usepackage{colortbl}
\usepackage{booktabs,siunitx}
 \usepackage{caption}
\usepackage{booktabs}
\usepackage{xcolor}
\usepackage[export]{adjustbox}

\usepackage{array}
\usepackage{amssymb}
\usepackage{pifont}
\def\KR#1{{\color{black} {{#1}}}} 

\usepackage{generic}
\usepackage{cite}
\usepackage{amsmath,amssymb,amsfonts}
\usepackage{algorithmic}
\usepackage{graphicx}
\usepackage{algorithm,algorithmic}
\usepackage{hyperref}
\usepackage{textcomp}
\def\BibTeX{{\rm B\kern-.05em{\sc i\kern-.025em b}\kern-.08em
    T\kern-.1667em\lower.7ex\hbox{E}\kern-.125emX}}
\def\KR#1{{\color{black} {{#1}}}} 
\def\KRA#1{{\color{black} {{#1}}}}
\begin{document}
\title{A Point Cloud Transformer for Remote Monitoring and Automated Assessment of Physical Rehabilitation Exercises}
\author{%
Kazi Rafat$^{1}$ \texttt{(kazi.meem@northsouth.edu)},\\
Md. Ismail Hossain$^{1}$ \texttt{(ismail.hossain2018@northsouth.edu)},\\
M M Lutfe Elahi$^{1}$ \texttt{(lutfe.elahi@northsouth.edu)},\\
Sifat Momen$^{1}$ \texttt{(sifat.momen@northsouth.edu)},\\
Fuad Rahman$^{2}$ \texttt{(fuad@apurbatech.com)},\\
Nabeel Mohammed$^{1}$ \texttt{(nabeel.mohammed@northsouth.edu)},\\
Shafin Rahman$^{1}$ \texttt{(shafin.rahman@northsouth.edu)}\\[4pt]
{\normalfont\small $^{1}$Department of Electrical and Computer Engineering, North South University, Dhaka, Bangladesh}\\
{\normalfont\small $^{2}$Apurba Technologies Limited, Dhaka, Bangladesh}\\}

\maketitle

\begin{abstract}

Rehabilitation exercises are essential in restoring lost physical functions of patients suffering from various diseases (e.g., Parkinson's, back pain). Carrying out these rehabilitation exercises, often prescribed by health experts, is costly, unavailable, and requires expert supervision. The availability of RGBD images and movement/position data of joints \KR{along with expert annotation of exercise data} has prompted the use of automatic assessment of the quality of rehabilitation exercises, which is cost-effective and can be carried out at home. However, existing approaches do not extract relevant features, lack practical application, require expensive pre-processing, or overlook crucial features. This study proposes a transformer-based framework for point clouds to extract features and assess rehabilitation exercises by analyzing joint positions collected through RGBD data. We adapt and utilize a curve-based point-cloud feature aggregation technique to augment point-cloud information that aids model output. The transformer architecture also uses axial self-attention, recognizing important joints and their roles to assist users in performing the exercise better. The guided system outperforms existing approaches and is also practically relevant due to its small size, fast inference, and generalization on specific joints in similar exercises. We conduct our experiments on three crucial baseline datasets for rehabilitation exercises: Kimore, UI-PRMD, and IRDS.

\end{abstract}


\section{Introduction}
\label{sec:introduction}

Physical therapy through rehabilitation exercises is effective in restoring lost functional abilities and rehabilitating patients suffering from different diseases (e.g., Parkinson's, back pain) that cause physical inability. Prescribed rehabilitation exercises are only powerful when performed correctly. Patients usually learn different rehabilitation exercises from physical therapists or informative videos on the Internet. Most patients (about 90\%)~\cite{komatireddy2014quality} practice these exercises at home without expert supervision. This affects exercise quality and effectiveness. Performing incorrect exercises can cause unwanted problems or further debilitate the patient. This also prevents patients from being motivated to continue or exercise regularly. In addition, physical therapy and rehabilitation appointments are costly and can put substantial financial burdens on patients and healthcare providers. The cost of physical rehabilitation programs in the US in 2007 was 13.5 billion dollars~\cite{machlin2011determinants}. \KR{The lack of trained rehabilitation professionals also increased the cost of rehabilitation programs.}

\begin{figure*}[!t]
  \centering
  \includegraphics[width=\linewidth]{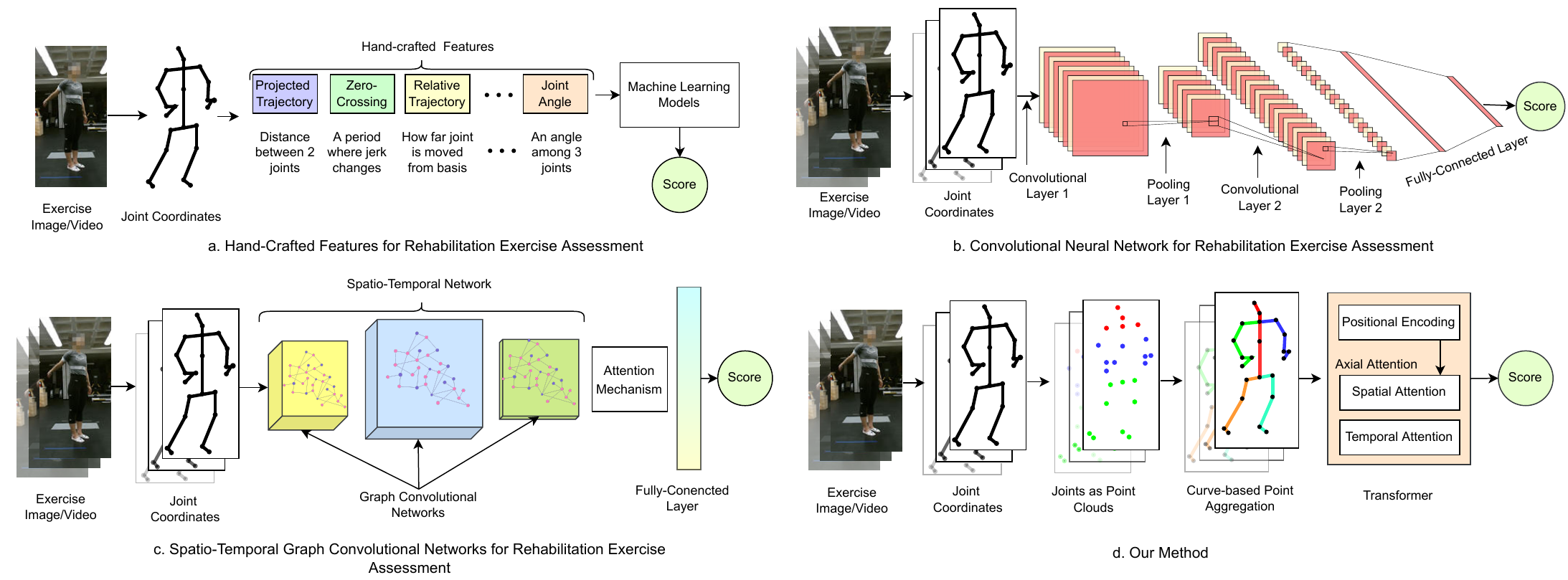}
\caption{We visually illustrate popular techniques against our technique. (a) Processing and Hand-crafting features~\cite{han2017space, yu2015discriminative, garcia2017transition
} require manual effort and introduce bias to the system. These techniques, at times, do not pay attention to important features. (b) Convolutional Neural Networks for assessing Rehabilitation~\cite{liao2020deep} exercises fail to capture some important spatio-temporal information. (c) Spatio-Temporal Graph Neural Networks~\cite{deb2022graph, bai2023improved} make topologically similar images and cannot map some topological information. (d) Curve-based cloud aggregation (\texttt{Ours}) paired with axial attention can augment data points according to their position and pay attention to important information.}
\label{fig:intro_diag}
\end{figure*}

\KR{Traditionally, patient progress in rehabilitation is assessed clinically through visual inspection, often based on comparisons with expert demonstrations or prior performances through qualitative annotations~\cite{howe2006community}. To reduce subjectivity and improve scalability, researchers have increasingly explored automated methods for assessing rehabilitation exercises. Early work~\cite{hamaguchi2020support, pogorelc2012automatic, yan2018spatial} primarily focused on classifying exercises as correct or incorrect. However, such binary classification provides limited feedback and does not support monitoring gradual improvements over time. More recent approaches instead formulate exercise assessment as a continuous evaluation problem, using regression models to assign performance scores across a range of quality levels~\cite{liao2020deep, deb2022graph}. These methods typically rely on movement and body-position data collected from sensors. In parallel, researchers have developed robotic assistive systems~\cite{maciejasz2014survey}, haptic devices~\cite{broeren2002virtual}, and virtual gaming environments~\cite{gauthier2017video} for exercise evaluation. However, such mechanical and electronic systems are often expensive and require specialized hardware, limiting their widespread adoption.}

Advancements in deep neural networks in classifying and interpreting vision tasks and their ease of use have made deep learning a frontrunner in the evaluation of physical rehabilitation tasks. The introduction of affordable motion and depth sensors, such as the Microsoft Kinect~\cite{zhang2012microsoft}, capable of tracking skeletal details, has spurred a growing interest in deep neural networks. These sensors can accurately record sequential joint positions or angles without lag. \KR{This high-frequency skeletal data allows for the differentiation between correctly executed exercise patterns and maladaptive exercise. In the context of rehabilitation, the ability to automatically distinguish these states allows for real-time feedback, ensuring that patients maintain the biomechanical alignment necessary for functional recovery and injury prevention, and allows the patient to continue and not leave physical therapy.} However, the comprehensive body data collected by motion sensors is often excessive and not ideal for direct use in neural networks. \KR{Kinect sensor captures nieghbouring data points in the same depth which have the same information making them spatially redundant, noisy or incomplete~\cite{song2020richly,wang2021human}. These redundancies result in an excessively large input space that is computationally expensive to process and prone to overfitting, as the model may focus on noise rather than the underlying biomechanical movement patterns.} Here, we summarize different strategies and discuss challenges in assessing exercise in physical rehabilitation.  
\textbf{(1)} Several studies~\cite{han2017space, yu2015discriminative, garcia2017transition} rely on handcrafted features that require experts to engineer features, which makes it costly, time-consuming and erroneous. \KR{Figure~\ref{fig:intro_diag}a shows hand-crafted features from joint coordinates which are then used to train machine learning models.}
\textbf{(2)} Studies~\cite{liao2020deep} also explored the use of computational deep learning, such as CNN, to extract features from joint positions in the body. The problem with CNNs as feature extractors is that they miss crucial spatial and temporal information in the skeleton data. \KR{Basic CNNs, Figure~\ref{fig:intro_diag}b, are unable to capture long-scale spatial and temporal information due to lack of attention mechanisms}
\textbf{(3)} Studies~\cite{deb2022graph},~\cite{bai2023improved} also used ST-GCNs (Spatio-Temporal Graph Convolution Networks) and LSTM to evaluate exercises. However, the graphs applied in ST-GCNs are predefined and are topologically similar across all frames that fail for different action classes and cannot map semantic information in all layers~\cite{shi2019two}. ST-GCNs cannot also map information such as bone lengths and directions, which are usually important. Existing approaches ignore joint importance or use self-attention to learn important joints, which is computationally inefficient~\cite{shi2019two}. Other studies~\cite{mourchid2023d, mourchid2023mr, shi2020skeleton} have also used modified GCNs with transformers and self-attention, \KR{but their performance was limited by the lack of ST-GCNs. Figure~\ref{fig:intro_diag}c shows st-gcns that are unable to map different topologies.}


In this study, we investigate the use of transformer models that utilize curve-based point cloud analysis~\cite{xiang2021walk} with axial attention \KR{to perform a regression task on skeletal sequences. Specifically, the model predicts a quantitative assessment score (normalized between 0 and 1) representing the kinematic similarity between the patient's performance and an expert's execution. This predicted score serves as a direct objective metric for tracking rehabilitation progress over time.} The curve-based point aggregation technique devices 'walks' on point clouds to form curves that are then aggregated to learn additional point features based on position and geometry. By employing relative encoding rules, the curve-based point aggregation method prevents points in the same geometry from having similar information as in other aggregation methods (local and non-local). This makes different points have distinguishable feature information and allows a neural network to learn and generalize better. Self-attention mechanisms require substantial calculations that are costly and time-consuming. As a result, we expand our model to employ axial attention that carries out the self-attention function on a rearranged video frame, enabling efficient calculation of attention coefficients. Attention coefficients determine the importance of a body joint in an exercise and allow users to focus on certain joints to enhance the quality of their exercise. We carry out experiments on baseline rehabilitation exercise datasets: Kinematic Assessment of Movement for Remote Monitoring of Physical Rehabilitation (KIMORE), University of Idaho-Physical Rehabilitation Movement Data (UI-PRMD), and IntelliRehabDS (IRDS) dataset. \KR{Figure~\ref{fig:intro_diag} showcases an illustrative comparison between past techinques against our method in Figure~\ref{fig:intro_diag}d.}

\KR{In this work, we propose a lightweight framework for automated rehabilitation exercise assessment that advances existing approaches in several ways. By incorporating a geometry-aware representation of skeletal joints, the model better preserves structural and kinematic relationships during movement. In addition, the use of axial self-attention enables efficient modeling of spatial–temporal dependencies across exercise sequences while maintaining low computational cost. Together, these design choices improve motion representation, scalability to variable-length sequences, and practical suitability for deployment in quasi real-world rehabilitation settings.}

\KR{Our proposed framework demonstrates competitive performance for automated assessment of rehabilitation exercises under standard benchmarking protocols. The main contributions of this work are summarized as follows:
\begin{itemize}
    \item We adapt a state-of-the-art curve-based point aggregation strategy utilizing geometry-aware, and variable-length modeling, originally introduced in CurveNet~\cite{xiang2021walk}, to skeletal joint point clouds, enabling the preservation of structural and kinematic information for exercise assessment.
    \item We integrate this representation with a transformer architecture employing axial self-attention to model spatial–temporal dependencies and estimate joint-level relevance during exercise execution.
    \item The proposed framework operates in an end-to-end manner and supports variable-length input sequences without requiring temporal alignment or segmentation.
    \item The proposed model is computationally lightweight compared to prior approaches, suggesting suitability for deployment in resource-constrained environments.
    \item Extensive experiments conducted on three publicly available benchmark datasets: KIMORE, UI-PRMD, and IRDS, demonstrate improved or competitive performance compared to existing state-of-the-art approaches.
\end{itemize}}


\section{Related Works}
\label{sec:related_works}

\subsection{Human movement and skeleton-based modeling}
\label{sec:related_works_1}

Human movement modeling has attracted the attention and interest of researchers in several domains. This has developed several subdomains that work on different techniques involving human movement and skeleton actions, as broad as human activity recognition~\cite{vrigkas2015review},~\cite{chen2012sensor},~\cite{ravi2005activity}, virtual reality~\cite{bower2015clinical} and into specifics such as human rehabilitation exercise classification~\cite{deb2022graph},~\cite{liao2020deep} and spontaneous smile classification~\cite{tahrim2022less}. Traditional methods of analysis and movement modeling resort to hand-crafted features~\cite{garcia2017transition},~\cite{yu2015discriminative},~\cite{xia2012view},~\cite{hamaguchi2020support},~\cite{pogorelc2012automatic} which heavily rely on modeling experts. \KR{Hand-crafted features tend to be limited as they are developed by experts thus introducing bias. Recently, hand-crafted features utilize neural networks~\cite{han2017space} as they are automatic and cost-effective.} Baccouche et al.~\cite{baccouche2011sequential} utilized CNN (Convolutional Neural Networks) for exercise classification where they used a 2-step procedure of CNN+RNN-T to access spatio-temporal understanding of the data. Lefebvre and colleagues~\cite{lefebvre2013blstm} used LSTMs (Long Short Term Memory) and RNNs (Recurrent Neural Networks) to classify gestures without any preprocessing step. LSTMs were also used by Ordonez and team~\cite{ordonez2016deep} for human activity recognition. \KR{LSTMs are able to learn important features but require a lot of data to learn or feature to feature intricacies due to no attention mechanisms. Along with that LSTMs are bad at removing redundancy and noise in data.} Spatio-Temporal graphs are also readily used to model and recognize human movements~\cite{jain2016structural},~\cite{deb2022graph} \KR{which are not able to capture all spatio-temporal information. At the same time ST-GCNs are not able to map all topological information.} Rahman et al.~\cite{rahman2022ai}  reported in a detailed review that deep learning techniques with attention outperform other techniques. Neural networks have also gained popularity due to recent developments in techniques that capture human motion quickly and effectively, such as Kinect~\cite{zhang2012microsoft}, Vicon~\cite{pfister2014comparative}, RGBD Cameras~\cite{das2011quantitative}, Accelerometers~\cite{troiano2014evolution} etc. \KR{However, such cameras either capture redundancy, inaccurate point positions or provide insufficient information.} Recent studies leverage graph convolutional networks (GCNs) to model spatio-temporal joint relationships in rehabilitation exercises, achieving real-time performance and high accuracy on clinical datasets like KIMORE and UI-PRMD \cite{kourbane2025optimized}. 
In this study, we investigate the provision of scores to assess rehabilitation exercises for a cost-effective and accurate home evaluation. \KR{The inconcistencies of image capturing techniques and the required preprocessing for the models to capture required features, we utilize a point-cloud aggregation technique that removes unnecessary information then augments points for better accuracy and feature understanding.}

\subsection{Exercise classification and quality assessment}
\label{sec:related_works_2}

Advancements in skeletal modeling have enabled prescribed exercise evaluation without the need for continuous professional supervision, which is often costly and time-consuming. \KR{Furthermore, performing rehabilitation exercises with incorrect posture over extended periods may negatively impact recovery. Typically, such systems compare a patient’s movements with those of healthy individuals performing the same exercises to determine correctness. These tasks are commonly formulated as classification problems using machine learning algorithms.}

\KR{Burns et al.~\cite{burns2018shoulder} employed machine learning techniques such as k-NN, SVM, and tree-based methods to evaluate shoulder physiotherapy exercises. Zhang et al.~\cite{zhang2011template} utilized a simplified k-NN approach with Euclidean distance for exercise classification through template matching. Crema et al.~\cite{crema2017imu} proposed an approach using inertial measurement units (IMUs) in commercial devices for automatic exercise detection and classification. Similarly, Lee et al.~\cite{lee2020automatic} applied conventional classification models to recognize squat postures. While these approaches achieve accurate classification, they primarily determine whether an exercise is correct or incorrect and do not provide continuous feedback on exercise quality.}

\KR{This limitation restricts a patient’s ability to understand incremental improvements or identify specific deficiencies in their movements. To address this, exercise quality assessment methods instead model the problem as a regression task, providing continuous scores to evaluate performance. Earlier approaches rely on distance-based measures~\cite{sakoe1978dynamic, houmanfar2014movement, hassan2008improving} to quantify similarity between exercises~\cite{su2014kinect, zhang2015objective, anton2015exercise}. However, such methods often require full sequence comparisons at each time step, which can limit their suitability for real-time feedback.}

\KR{More recent systems incorporate range-of-motion (ROM) classification and compensatory pattern recognition using deep learning, achieving 89–98\% accuracy in real-time evaluation~\cite{mennella2023deep}. Additionally, lightweight graph convolutional network (GCN)-based models provide continuous quality scores by learning joint relationships and temporal dependencies, typically requiring complete movement sequences for assessment~\cite{kourbane2025optimized}. Motivated by the effectiveness of deep learning approaches, we adopt a transformer-based model to enable real-time exercise quality assessment. }

\subsection{Deep Learning for Rehabilitation Exercise}
\label{sec:related_works_3}

In recent literature, approaches that follow deep learning methods to measure exercise quality have superseded other techniques. Zhang et al.~\cite{zhang2020semantics} used a semantics-guided neural network (SGN) for 3D skeleton-based action recognition and also introduced high-level semantics of joints into the neural network, utilized joint correlation, and utilized joint dependencies for enhanced model capability. Yan and colleagues~\cite{yan2018spatial} introduced the spatial-temporal graph convolutional network (ST-GCN) architecture that learns spatial and temporal patterns in images. Shahroudy et al.~\cite{shahroudy2016ntu} proposed a new recurrent neural network architecture that can utilize long-term temporal correlation of features, thus allowing the model to perform better. Du and his team~\cite{du2015hierarchical} brought about a new system of hierarchical recurrent neural networks that use subnets to model parts of the human skeleton, which are then fused into the hierarchy and fed into a fully connected network. However, little work has been done on rehabilitation exercises for different diseases. Recent work employs sequential GCNs to extract spatial features (joint relationships) and temporal features (frame correlations), outperforming state-of-the-art methods in rehabilitation assessment ~\cite{kourbane2025optimized}. Autonomous modeling using unsupervised phase learning and PCA anomaly detection has been proposed for stroke rehabilitation, achieving AUC scores of 0.9872 on compact motion capture systems ~\cite{jatesiktat2022autonomous}. Transformer-based architectures are increasingly adopted for real-time feedback, addressing limitations in traditional distance-based methods ~\cite{kourbane2025optimized,jatesiktat2022autonomous}. The assessment of physical rehabilitation exercises was first presented by Lee et al.~\cite{lee2019learning}, where they used hand-crafted features and a threshold model and classifiers to assess the quality of exercise. Liao et al.~\cite{liao2020deep} used spatio-temporal neural networks for automated assessment without manually crafting features. Recently, Deb et al.~\cite{deb2022graph} have thoroughly studied physical rehabilitation exercises. They used graph convolutions for physical rehabilitation and compared them to other impactful studies on skeleton-based data. In this study, we augment the information of 3D point clouds using curve-based cloud point analysis and learn sequential and spatial relations using a transformer-based architecture to better assess rehabilitation exercises.

\section{Methodology}
\label{sec:Methodology}

\subsection{Problem Formulation} 
\label{sec:problem_formulation}

\KR{Let a rehabilitation exercise dataset $\bm{X}$, such that $\bm{X} = \{\bm{x_1}, \bm{x_2}, \bm{x_3},.....,\bm{x_n}\}$, where $n \in \mathbb{Z}^+$ is the number of videos. Each video timestep $t$, consists of 3D skeletal joint information as point clouds $\bm{P}$, given as $\bm{P}\;=\{\bm{p_1}, \bm{p_2}, \bm{p_3},.....,\bm{p_m}\}$ where $\{\bm{p_i}:\bm{p_i} \in \mathbb{R}^3\}$, and $\{m:m\in \mathbb{Z}^+\}$. $p$ is a point in a point cloud, and $m$ is the number of joints present.}
\KR{Every timestep of video $x_i$, is superficially connected to its ground truth $y$, where $y=\{y_1, y_2, y_3,....., y_{b}\}$ is annotated such that $\{y_i: y_i\in \mathbb{R},[0,1]\}$ represents the quality of the performed exercise at time step $i$.}
\KR{This dataset is used to train a deep learning neural network, $s$, modeled as $\mathcal{F}_s(.,\bm{\mathcal{W}_s})$ where $\bm{\mathcal{W}_s}$ are model weights. The model $s$ is trained to produce outputs $y^\prime = \{y^{\prime}_1, y^{\prime}_2, y^{\prime}_3,....., y^{\prime}_{b}\}$ with $\{y^\prime_i:y^\prime_i \in \mathbb{R},[0,1]\}$. The training objective of the model is to minimize a loss function $\mathcal{L}(y, y^{\prime})$. This allows the neural network to generalize the exercise patterns to learn which action is beneficial and relevant to make the exercise successful. The exercise usually includes good and bad quality exercises carried out by a diverse group of experts and amateurs.}

\begin{figure*}[!t]
  \centering  
  \includegraphics[width=0.6\textheight]{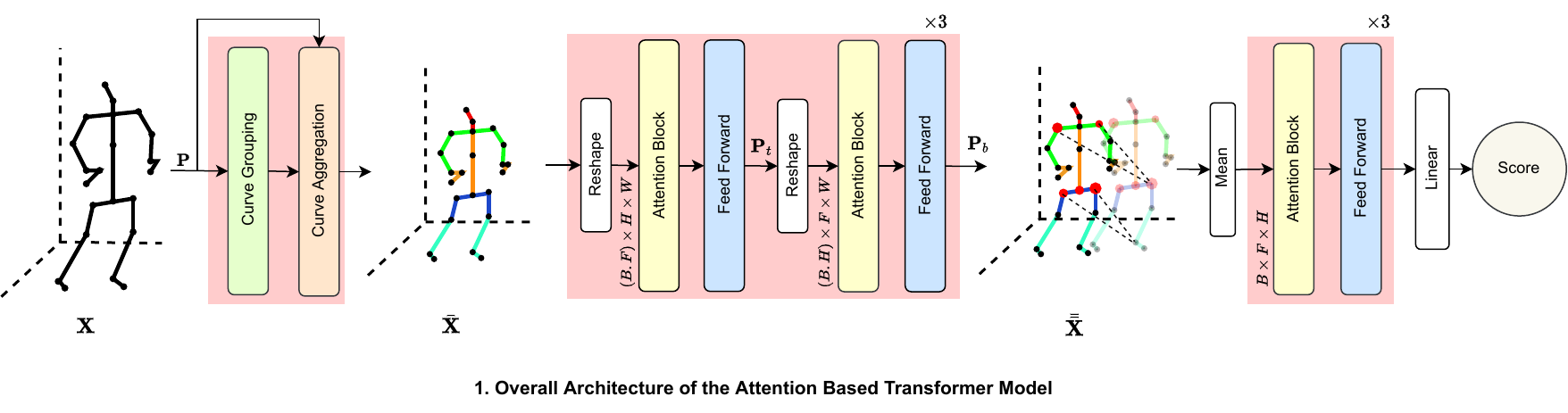}
    \includegraphics[width=0.6\textheight]{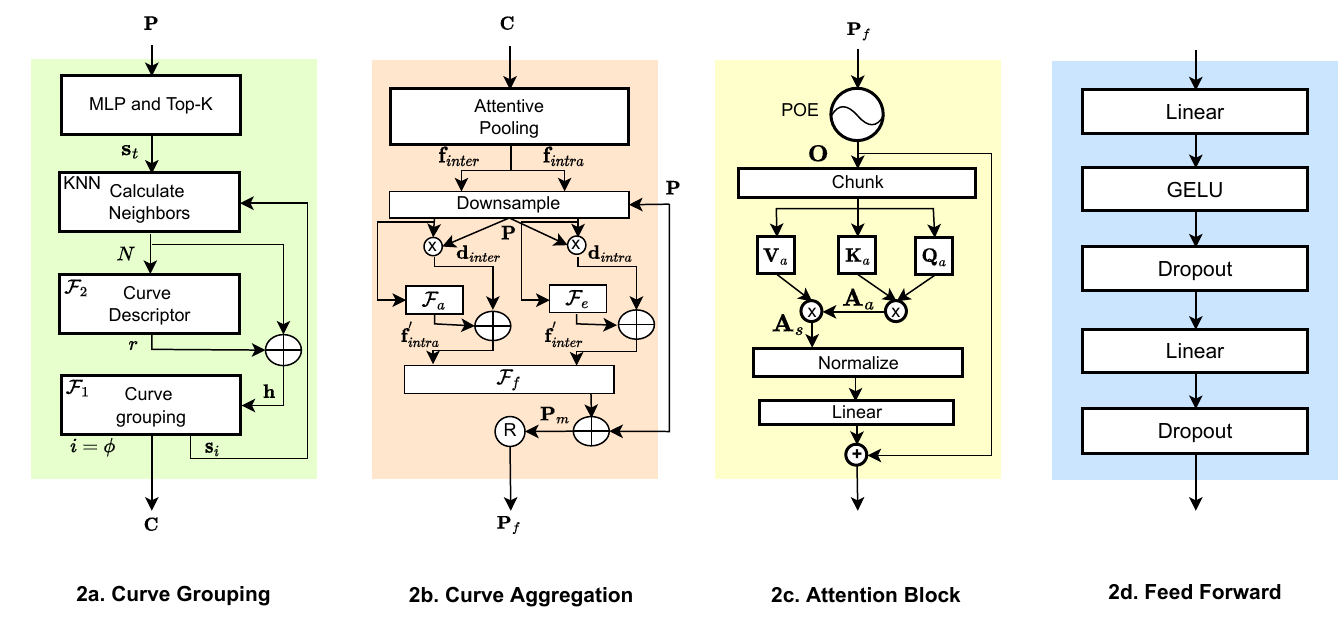}
\caption{
(1) \textbf{Architecture Overview:} Input $\bm{X}$ is passed through a curve-based point cloud aggregation block to produce $\bm{\Bar{X}}$, embedding spatial and shape features for each joint. $\bm{\Bar{X}}$ is then fed into a transformer with axial attention to model joint correlations across space and time, producing $\Bar{\Bar{\bm{X}}}$. A smaller attention block and a linear layer follow to compute the final score. The illustration shows a single curve's formation; multiple runs yield the full set of curves. 
(2a) \textbf{Curve Grouping:} Starting points $\bm{s_t}$ are selected from $\bm{P}$ using an MLP and Top-K. Neighbors $N$ are found via KNN. Curve descriptors $\bm{r}$ are learned via $\mathcal{F}_2$, concatenated with points to form $\bm{h}$, and processed by $\mathcal{F}_1$ to output curve points $\bm{s_i}$. 
(2b) \textbf{Curve Aggregation:} Learns inter-/intra-curve relations $\bm{f_{\text{inter}}}$, $\bm{f_{\text{intra}}}$, downsampled and combined into $\bm{f_{\text{inter}}'}$, $\bm{f_{\text{intra}}'}$, then passed through an MLP to get real points $\bm{P_f}$. 
(2c) \textbf{Attention Block:} Calculates position embeddings (POE), splits into $\bm{Q_a}$, $\bm{K_a}$, $\bm{V_a}$, and computes attention $\bm{A_s}$, added to POE for output. 
(2d) \textbf{Feed Forward Layer:} Restores the output to the original shape.
}

\label{fig:skel}
\end{figure*} 

\subsection{A Solution Framework and its Limitations}
\label{sec:solution_framework}

Most solutions consider evaluating rehabilitation exercises by extracting and relating the feature information of various points or body joints. Recent studies have looked at several feature aggregation techniques, such as local~\cite{wang2019dynamic} and non local operations~\cite{wang2018non}, multi-layer perceptrons (MLP)~\cite{qi2017pointnet},~\cite{qi2017pointnet++}, Graph Convolutional Networks (GCNs)~\cite{kipf2016semi}, and Spatio-temporal Graph Convolutional Networks (STGCN)~\cite{yu2017spatio} 

\KR{Local aggregation methods are particularly popular due to their effectiveness to learn relations and patterns among nearby points. Given a point $\bm{p_r} \in\bm{P}$ with neighbouring points $N$, such that $N = \{\bm{p_1}(r), \bm{p_2}(r), \bm{p_3}(r),.....,\bm{p_u}(r), N \in \bm{P}\}$ such that $u$ is the number of neighbors determined by KNN (K-nearest neighbors)~\cite{hu2020randla},~\cite{yan2020pointasnl} utilizing Manhattan or Euclidean distance. This distance, $\lambda$, between neighbors is stacked for aggregation and passed through MLPs (multilayer perceptrons), $\mathcal{F}_M(;\bm{\mathcal{W}_M})$ to obtain aggregated features $\bm{g}$. }

\KR{The issue with such encodings is that they are completely based on distance measurements. However, such distance-based encodings primarily capture local geometric proximity and may encode similar information for points lying on the same plane, particularly in shallow regions of the point cloud. Distinguishable feature representations tend to emerge mainly at structural boundaries, where point distributions differ more significantly. Furthermore, these approaches do not explicitly model temporal interdependencies between joints across time, nor do they account for the relative importance of individual joints within a single frame. As a result, joint-level relationships critical for assessing exercise quality may be insufficiently captured, limiting the effectiveness of these methods in providing actionable feedback for rehabilitation.}

\KR{Recent popular solutions use self-attention~\cite{vaswani2017attention} mechanisms for movement modeling, to relate different spatial regions within a single input to one another. An embedding representation of a single frame instance $\bm{P} = \{\bm{p_1}, \bm{p_2}, \bm{p_3},....., \bm{p_m}\}$ where each joint embedding, $\bm{p_i}$, is projected using three learnable linear transformations parameterized by $\bm{\mathcal{W}_Q}, \bm{\mathcal{W}_K},$ and $\bm{\mathcal{W}_V}$. Query ($\bm{Q}$), key ($\bm{K}$) and value ($\bm{V}$) formulated as $\bm{Q} = \mathcal{F}_Q(\bm{P};\bm{\mathcal{W}_Q})$, $\bm{K} = \mathcal{F}_K(\bm{P};\bm{\mathcal{W}_K})$, and $\bm{V} = \mathcal{F}_V(\bm{P};\bm{\mathcal{W}_V})$ such that the matrices produced by these functions have a dimentionality $d$. The query and key are calculated across all joints, and the similarity between $\bm{Q}$ and $\bm{K}$ is measured using a scaled dot-product similarity. The similarity measures the relation between the query joint and the key joint. This produces the attention filter $\bm{A}$ given by $\bm{A} = softmax(\frac{\bm{Q}.\bm{K}^T}{\sqrt{d}})$, which, after training, tends to produce higher values for similar joints. The attention filter $\bm{A}$ is then applied to $\bm{V}$ to get only the important feature relations. The self-attention mechanism can be represented by the equation $Attention(\bm{Q},\bm{K},\bm{V})=\bm{A}.\bm{V}=softmax(\frac{\bm{Q}.\bm{K}^T}{\sqrt{d}}).\bm{V}$. However, when considering multi-dimensional data, self-attention techniques are inefficient as they take $O(N^2)$, where $N$ denotes the number of input tokens (e.g., joints or spatial locations). Multi-dimensional inputs, even for relatively small inputs $3 \times 24 \times 24$, when flattened, have a size of $1728$ and are too long to find relations between points.}

\subsection{Curve Based Transformer with Joint Attention}
\label{sec:Curve_Based_Transformer}

We investigate the SOTA aggregation technique of curve-based point cloud aggregation~\cite{xiang2021walk} that looks at feature aggregation by grouping curves through landmarks (joints) for a better overall depiction of a point cloud. The curve aggregation technique operates on two different hyperparameters, the curve number ($\eta$) and the curve points ($\phi$). To group $\eta$ curves, determining the starting point of a curve $c$ is important and has to be done beforehand. \KR{To determine the starting point, the Top-K selection~\cite{gao2019graph} method is used along with a multi-layer perceptron among all points $\bm{P}$. From the starting point $\bm{s_1}\in\bm{P}$, a walk $W$, groups $\phi$ points, using a walk policy $\pi$, for a curve $c$, containing points $S$, where $S = \{\bm{s_1}, \bm{s_2}, \bm{s_k},\bm{s_{k+1}},.....,\bm{s_{\phi}}\}$.} We determine the next point in the curve using the walk policy ($\pi$): $\bm{s_{k+1}} = \pi(\bm{s_k})$ where $k\in \mathbb{Z^+},\;k\leq\phi$. \KR{To create the walk policy, a state descriptor $\bm{h^{s_{i}}_{N}}$, where $\bm{h^{s_{i}}_{N}}\in\mathbb{R}^{2||s||}$ is created through the concatenation of the neighboring points $N$, of point $\bm{s_i}$, such that $N = \{\bm{n^1},\bm{n^2},.....,\bm{n^j}\}$ and the curve descriptor $\bm{r_i}$ (discussed later) to give $\bm{h^{s_{i}}_{N}}$, on which an MLP, given as function $\mathcal{F}_1$, is applied to learn scores $\alpha$.} The state descriptor with the best neighbor gives the highest score and is selected for the curve. So the walk policy can be formulated as:
\begin{equation}
    \bm{h^{s_{i}}_{N}} = N \bigoplus \bm{r_i} 
\end{equation}
\begin{equation}
    \alpha = {\mathcal{F}_1(\bm{h^{s_{i}}_{N}})}  
\end{equation}
\begin{equation}
     \pi(\bm{s_i}) = argmax(softmax(\alpha))
\end{equation}


A Gumbel-softmax~\cite{jang2016categorical} probability is usually sought after, as the \textit{argmax} function usually reduces the gradient. \KR{The curve is extended towards $\bm{n^i}$ (the best neighbour)$ = \bm{s_{k+1}} = \pi(\bm{s_k})$. To avoid curve loops, a point, $s_i$ is concatenated with a curve descriptor.} The curve descriptor, $\bm{r_i}$ for every point $\bm{s_i}$, where $\bm{r_i} \in \mathbb{R}^{||s||}$, is a trainable variable that encodes the state of the curve at a certain step. Dynamic momentum is used to calculate the value of a variable $\beta$ using an MLP, $\mathcal{F}_2$, to learn $\bm{r_i}$.
\begin{equation}
    \beta = softmax(\mathcal{F}_2(\bm{r_{i-1}}, \bm{s_i}))
\end{equation}
\begin{equation}
    \bm{r_i} = \beta \bm{r_{i-1}}+(1-\beta) \bm{s_i}
\end{equation}

Even so, crossovers or the curve intersecting itself might be problematic even if loops are avoided. Many crossovers suggest particular points to be encoded several times, which may hurt the point-cloud representation. \KR{To represent the direction of the curve at any point $i$ in the walk, a vector $\overrightarrow{\bm{c_i}} = \bm{s_i}- \bm{r_{i-1}}$ is created.} \KR{Similarly, a vector $\overrightarrow{\bm{q_i^j}} = \bm{n^j}-\bm{s_i}$ can be calculated for every neighbour for the point $\bm{s_i}$.} The angle $\theta$, between the vectors determines whether the curve will move to a point at an angle or a straight line, which is calculated using cosine similarity. The higher the value of $\theta$, the higher the chances of cross-overs. So, crossovers are suppressed by reducing the value of $\alpha^{j}$ for neighbors $\bm{n^j}$ that have a higher $\theta$.

After completing the $\eta$ curves, $C\in\mathbb{R}^{C\times\eta \times\phi}$, $C= \{c_1, c_2, c_3,....., c_{\eta}\}$, the curve aggregation technique looks into the relations between the curves and the curve points represented as vectors $\bm{f_{inter}} \in \mathbb{R}^{C\times\eta}$ and $\bm{f_{intra}} \in \mathbb{R}^{C\times\phi}$, respectively. These vector representations are learned using an Attentive-Pooling~\cite{hu2020randla} operator, once along a curve and once with all the curves by altering the pooling dimensions. \KR{The curve relations $\bm{f_{inter}}$ and $\bm{f_{intra}}$ are downsampled to reduce unnecessary features.} The downsampled curve relations are multiplied by the downsampled points separately to learn the curve mappings to get $\bm{d_{inter}}$ and $\bm{d_{intra}}$. Vectors $\bm{f_{intra}}$ and $\bm{f_{inter}}$ are further processed using two MLPs, $\mathcal{F}_e$ and $\mathcal{F}_a$ and are fused with curve mappings to get $\bm{f'_{inter}}$ and $\bm{f'_{intra}}$. The curve mappings $\bm{f'_{inter}}$ and $\bm{f'_{intra}}$ are then passed through another MLP, $\mathcal{F}_f$. Finally, these mappings are concatenated with the points $\bm{P}$ to give $\bm{P_m}$. This output is then used as a residual addition to the original input to get $\bm{P_f}$.

\begin{table*}[!t]
\centering
\setlength\heavyrulewidth{0.3ex}

\sisetup{table-format=6.0}
\caption{Description of UI-PRMD, Kimore, and IRDS dataset.}

\begin{tabular}{cccccccccc} 
\toprule

  \multirow{2}{*}{Dataset}
  &\multirow{2}{*}{Exercise} &  \multirow{2}{*}{Patients} & \multirow{2}{*}{Participants} &\multirow{2}{*}{Sensor} &\multirow{2}{*}{Wearable} &  No. of  & No. of & \multirow{2}{*}{Annotators} & \multirow{2}{*}{Type}\\
& & & & & &Exercise &Joints & & \\ \cmidrule[\heavyrulewidth](){1-10}

\multirow{2}{*}{UI-PRMD~\cite{vakanski2018data}}  & Physical Therapy  &\multirow{2}{*}{0}  &\multirow{2}{*}{10}  & \multirow{2}{*}{KinectIJP-JO}   & \multirow{2}{*}{Vicon} & \multirow{2}{*}{10} & \multirow{2}{*}{39 (Vicon)} & \multirow{2}{*}{Instructions} & \multirow{2}{*}{Assessment}\\
& and Rehabilitation & &&&&&&&\\
\arrayrulecolor{lightgray}
\midrule


 \multirow{2}{*}{Kimore~\cite{capecci2019kimore}} & Posture and Back- & \multirow{2}{*}{34} & \multirow{2}{*}{78} & \multirow{2}{*}{Kinect2RDB} & \multirow{2}{*}{NA} & \multirow{2}{*}{5} & \multirow{2}{*}{25} & \multirow{2}{*}{3 Meds}&\multirow{2}{*}{Assessment} \\
 &pain Rehabilitation& &&&&&&&\\
 \midrule
 \multirow{2}{*}{IRDS~\cite{miron2021intellirehabds}} & General Physical & \multirow{2}{*}{15} & \multirow{2}{*}{29} & \multirow{2}{*}{Kinect2JP} & \multirow{2}{*}{NA} & \multirow{2}{*}{9} & \multirow{2}{*}{29} & \multirow{2}{*}{2 Non-Meds}&\multirow{2}{*}{Classification} \\
 &Rehabilitation& &&&&&&&\\ 
 \arrayrulecolor{black}

 \bottomrule

\end{tabular}

    \label{tab:Datasets}
\end{table*}

The curve-based aggregation of point clouds is downsampled using a convolution layer to reduce the necessary calculations that is required. The inputs $\bm{P_f}$ are then normalized and passed to a transformer layer that utilizes a quite similar attention mechanism like axial attention~\cite{wang2020axial},~\cite{ho2019axial} to process the data to have attention across different dimensions. Considering an input exercise video $\bm{x_i}$, with shape $B\times F\times H\times W$, consists of $H$ landmarks with $W$ information for that landmark, for $F$ frames sampled at a batch size $B$, is rearranged so that attention along different axis/dimension can be calculated while information along the other dimension is independent. \KR{We consider the landmarks to be the agumented cloud points $\bm{P_f}$ that is passed on to the transformer layer. $W$ represents the coordinates (available information in the landmark.)} Firstly, the input $B\times F\times H\times W$ is rearranged in the shape $(B.H)\times F \times W$. This arrangement allows for calculating the self-attention of a landmark with other landmarks in all the other frames of its timestep. Attention is calculated similarly to the original self-attention mechanism. The position embedding, $\bm{O}$ of every data point is calculated and separated into three chunks assigned as Query ($\bm{Q_a}$), Key ($\bm{K_a}$), and Value ($\bm{V_a}$) such that these matrices have a size of $d_a$. Attention filters $\bm{A_a}$ can be calculated where $\bm{A_a}  = softmax(\frac{\bm{Q_a}.\bm{K_a}^T}{d_a})$. Self-attention can be calculated by multiplying the filter by its value such that $Axial\;Self\;Attention(\bm{Q_a}, \bm{K_a}, \bm{V_a}) = \bm{A_a}.\bm{V_a} = softmax(\frac{\bm{Q_a}.\bm{K_a}^T}{d_a}).\bm{V_a}$. This axial self-attention matrix, $\bm{A_s}$, is passed through a linear layer and is added to the rearranged input tensor to obtain a tensor, $P_t$, with attention in the time domain. After this, the rearranged input tensor is passed through a linear layer $\mathcal{F}_t$ with weights $W_a$. Finally, this output is added to the rearranged input embedding. After that, the input tensor is rearranged in its original shape, $B\times F\times H\times W$. It is rearranged again in the shape $(B.F)\times H \times W$ so that the self-attention of every reference point can be calculated with other reference points in one frame. As a result, the output tensor is oriented toward landmarks in the same frame and different time frames represented by $P_b$. Similarly, attention is calculated and passed through another linear layer. The input tensor is reshaped to its original position. This is done 3 times, considering that there are 3 chunks. The data point information of the output tensor from this multi-attention layer is averaged and passed onto another multi-layer attention layer. This layer is similar to the previous attention mechanism, but the input tensor does not require rearrangement. Finally, the outputs of the second attention layer are passed through a linear layer to get scores.


\section{Experiments}
\label{sec:Experiments}

\subsection{Setup}
\label{sec:Setup}
\noindent\textbf{Datasets:} We evaluate our system on three physical rehabilitation datasets: \textbf{\textit{(1)} KIMORE}~\cite{capecci2019kimore} contains RGB, depth videos and skeleton data for 5 lower back pain exercises performed by 78 subjects (44 healthy controls including 12 physiotherapists, 34 patients with chronic motor diseases). Exercises include trunk stretching, arm extension, trunk rotation, pelvic rotation, and squatting, with physician-annotated scores and 25 joint positions captured. \textbf{\textit{(2)} UI-PRMD}~\cite{vakanski2018data} consists of 10 healthy subjects performing 10 repetitions each of 10 exercises (deep squat, hurdle step, lunges, sit-to-stand, shoulder movements) captured via Microsoft Kinect (22 joints) and Vicon optical tracker (39 joints). \textbf{\textit{(3)} IntelliRehabDS}~\cite{miron2021intellirehabds} includes 9 rehabilitation exercises performed by 29 individuals (15 patients) with gestures classified as correct or incorrect, captured using 25 Kinect joint positions per frame.
\noindent\textbf{Evaluation Metric:} We evaluate model performance using three standard metrics: Mean Absolute Deviation (MAD), Mean Absolute Percentage Error (MAPE), and Root Mean Squared Error (RMSE). Lower values indicate better performance. The metrics are defined as:
\begin{equation}
    \text{MAD}(y, \hat{y}) = \frac{1}{N} \sum_{i=0}^{N - 1} |{y_i - \hat{y}_i}|.
    \label{eqn:MAD}
\end{equation}
\begin{equation}
    \text{MAPE}(y, \hat{y}) = \frac{100}{N} \sum_{i=0}^{N - 1} \frac{|y_i - \hat{y}_i|}{y_i}.
    \label{eqn:MAPE}
\end{equation}
\begin{equation}
    \text{RMSE}(y, \hat{y}) = \sqrt{\frac{1}{N} \sum_{i=0}^{N - 1} {
    (y_i - \hat{y}_i)^2}}.
    \label{eqn:RMSE}
\end{equation}

\KR{\noindent\textbf{Architectural Hyper-parameter Validation:} We split our dataset into three sets on a ratio 80:10:10 for training testing and validation. We make sure the test set does not contain instances of patients that can be found in the train or validation set. All the hyper-parameter validations are done per dataset basis on the validation set, because the hyper-parameters of a curve-aggregation architectures is based on the number of joints. We carry out the hyper-parameter ablation on exercise 5 of the KIMORE dataset and exercise 10 of the UI-PRMD dataset.  We performed hyperparameter validation on the number of neighbors, curve length and curve quantity which have a crucial effect on the model's performance in predicting exercise quality.} \KRA{We treat the number of neighbours $j$ used during the curve-walking procedure $\pi$ as a hyperparameter and evaluate the integer range $j \in \{4,\dots,10\}$. The lower bound $j=4$ is the minimum required to define a stable local surface orientation in $3$D space. Above $j=10$, the neighbourhood spans multiple anatomical segments and aggregates unrelated joints, which degrades localisation and raises computational cost. Across all datasets, the lowest validation error was obtained at $j=9$; we therefore adopt the neighbour set $\mathcal{N} = \{\bm{n^1}, \bm{n^2}, \dots, \bm{n^j}\}$ with $|\mathcal{N}| = j = 9$ for all experiments.}
\KR{The curve-based aggregation architecture utilizes important hyperparameters such as curve length and curve quantity. Short curves cannot capture and aggregate patterns from all important neighbors. Longer curves may capture unnecessary and non-correlating information and are computationally expensive. We tune the curve number and the curve length for each exercise to understand which combination of curve length and curve number can properly capture important information during aggregation. We conduct tuning experiments on a wide range of curve lengths and quantities. We visualize a 3D surface plot, in Figure~\ref{fig:Ablation_points},of combinations of the number of points and curves against the Mean Absolute Error \KR{for the validation set of} Kimore Ex 5. We compare all combinations of $10, 15, 20$ points with $3, 5, 8$ curves for the Kimore ex 5. For the KIMORE dataset, a combination of 5 curves with 15 points can best predict exercise quality on the validation set. For the UI-PRMD dataset, 5 curves with 25 points, can best predict exercise quality among all combinations of $15, 20, 25, 30$ points with $3, 5, 8$ curves. With this, we can conclude the proper length and number of curves required for each exercise to obtain only essential information so that the model performs accurately and is computationally efficient.}

\KR{We do not do hyper-parameter ablation on the IRDS dataset and consider the number of curves and points to be 5 and 15 respectively.}

\noindent\textbf{Implementation details\footnote{Code and models are available at Github link: \url{https://github.com/King-Rafat/Transformer_Rehabilitation}}} The model for our experiments was trained using an Adam optimizer for 2000 epochs with a learning rate of $1e^{-4}$ using a Huber loss~\cite{meyer2021alternative} for backpropagation. We use two blocks containing a curve aggregation and a curve grouping block for the curve-based feature aggregation, unlike its paper, which utilizes 4 blocks. We measured the performance metrics of the model on the test set after every epoch to get the optimal model. We used the experimental scores, which are reported as the average of 10 repetitions for each exercise in each data set. We used the PyTorch (v1.13 and CUDA 11.3) deep learning framework for the experiments. The experiments were conducted on a workstation with a single NVIDIA RTX 3090 GPU and 32GB of RAM.

\begin{table*}[!t]
\rowcolors{4}{gray!30}{white} 
\begin{minipage}{\columnwidth}
\caption{Results on the UI-PRMD dataset showcasing the evaluation metric MAD on \texttt{Our} technique and comparisons to other techniques. Lower values indicate better results.}
\begin{tabular}{c |c| c c c |cc}
\toprule
 & & \multicolumn{3}{c}{\KRA{GCN-based}} & \multicolumn{2}{|c}{\KRA{CNN-based}}\\
\toprule
 \multirow{2}{*}{Ex} & \multirow{2}{*}{\texttt{Ours}} &  Mourchid  & Deb & Song   & Liao  & \KRA{Deep}\\
& & et al.~\cite{mourchid2023mr}&et al.~\cite{deb2022graph} & et al.~\cite{song2020richly}  & et al.~\cite{liao2020deep} & \KRA{CNN}\\
\toprule
1 &  \textbf{0.008}& 0.008  & 0.009 & 0.011  & 0.011 &  \KRA{0.014}\\ 
 2 & \textbf{0.005}& 0.010 &  0.006 & 0.006  & 0.028 & \KRA{0.029}\\
3 & \textbf{0.006}& 0.010  & 0.013 & 0.010  & 0.039 & \KRA{0.041}\\
 4 & \textbf{0.006}& 0.008 & 0.006 & 0.014  & 0.012 & \KRA{0.016}\\
 5 & \textbf{0.006} &0.007 & 0.008 & 0.013  & 0.019 & \KRA{0.013}\\
 6 & \textbf{0.006} &0.010& 0.006 & 0.009 & 0.018 & \KRA{0.023} \\
 7 & \textbf{0.010} &0.020&  0.011 & 0.017 & 0.038 & \KRA{0.033}\\
 8 & \textbf{0.014} &0.020&  0.016 & 0.017 &0.023 & \KRA{0.029} \\
 9 & \textbf{0.008} &0.014&  0.008 & 0.008  & 0.023 & \KRA{0.025} \\
 10 & \textbf{0.018} &0.015&  0.031 & 0.038  & 0.042 & \KRA{0.037} \\
 \bottomrule
\end{tabular}

\label{tab:UI-PRMD}
\end{minipage}
\hspace{0.5cm}
\rowcolors{4}{gray!30}{white}
\begin{minipage}{\columnwidth}
\caption{Results on the IRDS dataset showcasing the evaluation metric accuracy on \texttt{Our} technique and comparisons to other techniques. Higher values indicate better results.}
\tabcolsep=0.17cm
\begin{tabular}{c |c| c c |cc}
\toprule
 & & \multicolumn{2}{c}{\KRA{GCN-based}} & \multicolumn{2}{|c}{\KRA{CNN-based}}\\
\toprule
 \multirow{2}{*}{Ex} & \multirow{2}{*}{\texttt{Ours}} &  Zheng et al. & Zheng et al. & Zhang & Li  \\
& & Baseline.~\cite{zheng2023skeleton}& R.I.~\cite{zheng2023skeleton}&et al.~\cite{zhang2019view} & et al.~\cite{li2020learning}\\
 \toprule
 1 &  \textbf{1.0000}& 0.9697 & 0.9697 & 0.9697 & 0.9848 \\ 
 2 & \textbf{0.9571}& \textbf{0.9571} & \textbf{0.9571}&  \textbf{0.9571} & 0.9429 \\
 3 & 0.9787& \textbf{0.9894} & 0.9681 & 0.9681 & 0.9787 \\
 4 & \textbf{0.9740}& \textbf{0.9740} & \textbf{0.9740} & \textbf{0.9740} & \textbf{0.9740} \\
 5 & 0.9857 &0.9714  & 0.9857 & 0.9714 & \textbf{1.0000} \\
 6 & \textbf{0.9848} &\textbf{0.9848} & \textbf{0.9848} & \textbf{0.9848} & \textbf{0.9848} \\
 7 & \textbf{1.0000} & \textbf{1.0000} & \textbf{1.0000} &  \textbf{1.0000} & 0.9683\\
 8 & \textbf{0.9706} &\textbf{0.9706} & 0.9412&  0.9412 & 0.9559\\
 9 & \textbf{0.9863} &0.9589 & \textbf{0.9863}&  0.9452 & 0.9589\\
\textbf{Mean} & \textbf{0.9819} &0.9751& 0.9741& 0.9680 &0.9720 \\
 \bottomrule 
\end{tabular}
\label{tab:IRDS}
\end{minipage}
\end{table*}

\begin{table*}[!t]
\tabcolsep=0.195cm
\begin{minipage}[t]{0.66\textwidth}
\vspace{-3.2cm}
\caption{Experimental Results on the KIMORE dataset showcasing the evaluation metric MAD, MAPE, and RMSE on \texttt{Our} technique and comparisons to other techniques. Lower values indicate better results.}
\rowcolors{4}{gray!30}{white}
\begin{tabular}{c |c| c | c c c c |c c}
\toprule
 & & & \multicolumn{4}{c}{\KRA{GCN-based}} & \multicolumn{2}{|c}{\KRA{CNN-based}}\\
 \toprule
 \multirow{2}{*}{Metric} & \multirow{2}{*}{Ex} & \multirow{2}{*}{\texttt{Ours}}& Mourchid & Deb & Song  & Yan &  Liao & \KRA{Deep}\\
& & & et al.~\cite{mourchid2023d}&et al.~\cite{deb2022graph} & et al.~\cite{song2020richly} & et al.~\cite{yan2018spatial} & et al.~\cite{liao2020deep} & \KRA{CNN}\\
 \midrule
  \cellcolor{white}\multirow{-5}{*}{} & 1 & \textbf{0.185}  & 0.641& 0.799 & 0.977 & 0.889  &  1.141 & \KRA{1.445}\\ 
 \cellcolor{white}& 2 & \textbf{0.560} & 0.753& 0.774 & 1.282 & 2.096 & 1.528 & \KRA{1.768}\\
\cellcolor{white}MAD & 3 & \textbf{0.128} & 0.210& 0.369 & 1.105 & 0.604  &  0.845 & \KRA{1.115} \\
\cellcolor{white} & 4 & 0.256 & \textbf{0.206} & 0.347  & 0.715 & 0.842  &  0.468 & \KRA{1.001} \\
\cellcolor{white} & 5 & \textbf{0.388}  & 0.399& 0.621 & 1.536 & 1.128   &  0.847 & \KRA{1.672} \\
 \midrule
\cellcolor{white} \multirow{-5}{*}{} & 1 & \textbf{0.591}  & 2.020 & 2.024 & 2.165 & 2.017 &  2.534 & \KRA{2.156}\\ 
\cellcolor{white}& 2 & \textbf{1.235}  & 1.468 & 2.120 & 3.345 & 3.262 &   3.738 &  \KRA{3.899} \\
\cellcolor{white}RMSE& 3 & \textbf{0.233}  & 0.487 & 0.556 & 1.193 & 0.799  & 1.561 & \KRA{1.345} \\
\cellcolor{white}& 4 & \textbf{0.451}  & 0.527 & 0.644  & 2.018 & 1.331 & 0.792 & \KRA{2.135}\\
\cellcolor{white} & 5 & \textbf{0.678}  & 0.735  & 1.181 & 3.198 & 1.951 & 1.914 & \KRA{1.973} \\
 \midrule
\cellcolor{white} \multirow{-5}{*}{} & 1 &  \textbf{0.543} & 1.623 & 1.926 & 2.605 & 2.339 &  2.589 & \KRA{2.872}\\ 
\cellcolor{white}& 2 & \textbf{1.891}  & 0.974 & 1.272 & 3.296 & 6.136 & 3.976  & \KRA{4.182}\\
\cellcolor{white}MAPE& 3 & \textbf{0.336}  & 0.613 & 0.728 & 2.968 & 1.727 & 2.023 & \KRA{2.221}\\
\cellcolor{white}& 4 & 0.766& \textbf{0.541} & 0.824  & 2.152 & 2.325 &  2.333 & \KRA{2.641}\\
\cellcolor{white}& 5 & \textbf{1.199} & 1.217  & 1.591 & 4.959 & 3.802 &  2.312 & \KRA{2.021} \\
 \arrayrulecolor{black}
 \bottomrule
\end{tabular}
\label{tab:Kimore}
\end{minipage}
\hspace{0.2cm}
\begin{minipage}[t]{0.3\textwidth}
\includegraphics[height = 3.5cm, width =6cm,  angle=-0.4]{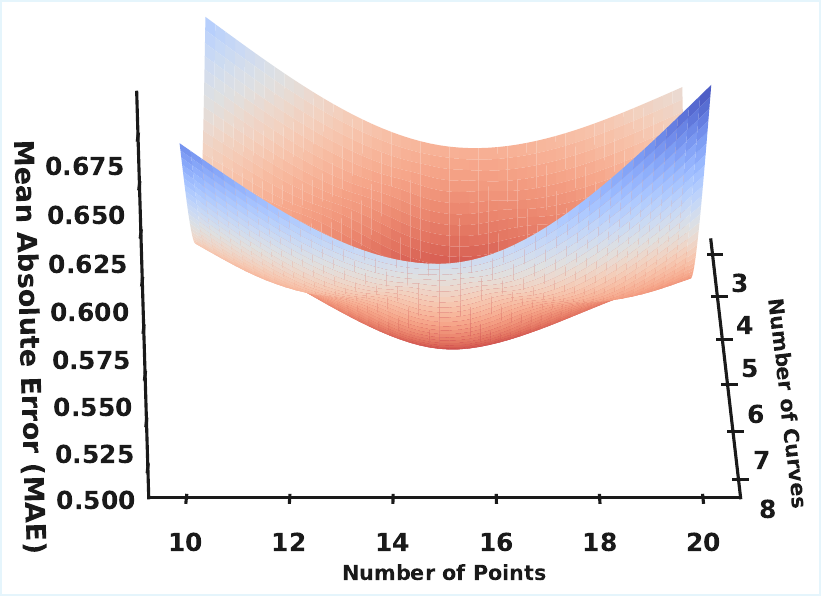}
\captionof{figure}{Points and Curves against Mean Absolute Error for Kimore Ex 5}
\label{fig:Ablation_points}
\includegraphics[scale = 0.15]{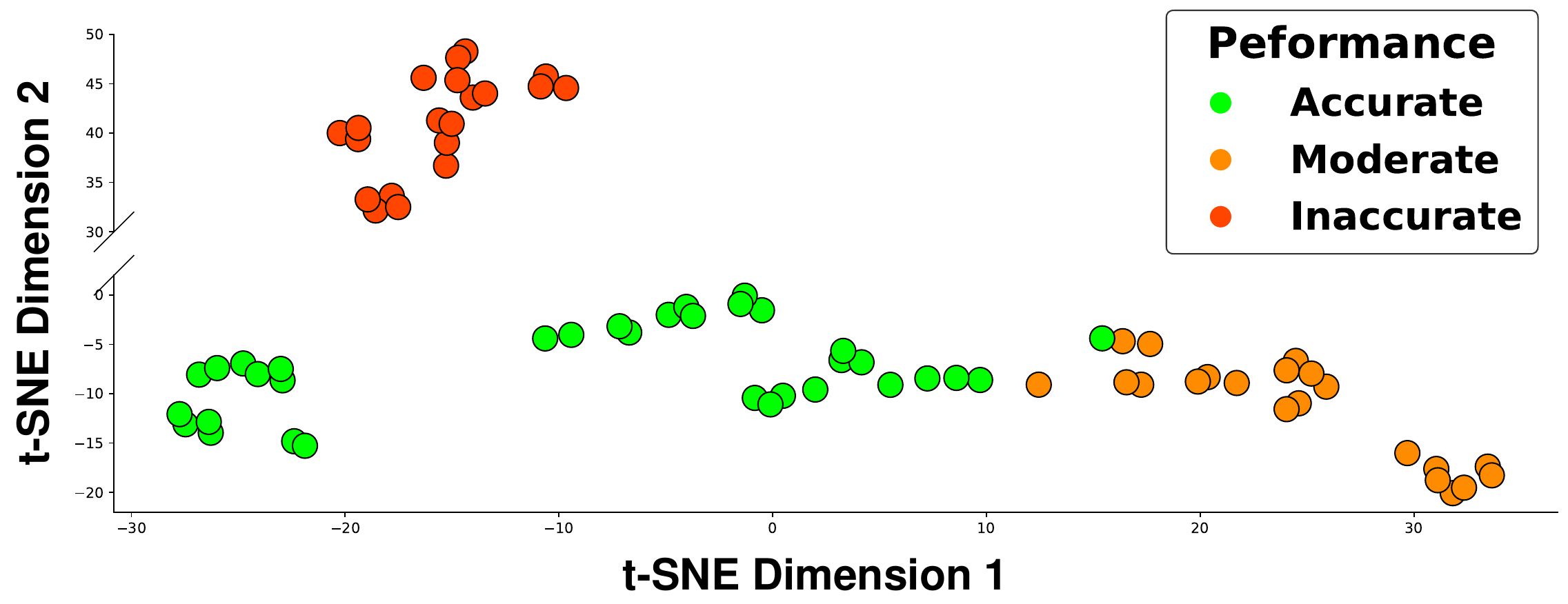}
\caption{t-SNE visualization show distinct clusters between accurate, moderate, and inaccurate exercises.}
\label{fig:TSNE}
        
\end{minipage}
\end{table*}

 



\begin{table*}[!t]
\begin{minipage}[t]{0.73\textwidth}
\caption{Ablation Study on the UI-PRMD dataset showcasing the evaluation metric MAD. CVN refers to the curve-based aggregation module. TL is the Transformer Layers. Lower values indicate better results.}
\begin{tabular}{p{0.5cm} p{0.5cm} p{0.5cm} c c c c c c c c c c} 
 \toprule
 CVN & TL1 & TL2 & Ex 1 & Ex 2 & Ex 3 & Ex 4 & Ex 5 & Ex 6 & Ex 7 & Ex 8 & Ex 9 & Ex 10\\
\midrule 
Yes &  Yes & Yes &\textbf{0.008} & \textbf{0.005} & \textbf{0.006} & \textbf{0.006} & \textbf{0.006} & \textbf{0.006} & \textbf{0.010} & 0.014 & \textbf{0.008} & \textbf{0.018}\\ 
Yes &  No & Yes &0.010 & 0.009 & 0.009 & 0.008 & 0.008 & 0.007 & 0.010 & \textbf{0.011} & 0.010 & 0.044\\ 
Yes & Yes & No & 0.011 & 0.008 & 0.009& 0.009& 0.010 & 0.007 & 0.009& \textbf{0.011} & 0.009&0.041\\
 No & Yes & Yes & 0.016& 0.012& 0.012 & 0.013 &0.016& 0.013 & 0.010& 0.014& 0.017 & 0.028\\
 \bottomrule
\end{tabular}

    \label{tab:UI-PRMD-Abl}
\end{minipage}
\hspace*{0.2cm}
\begin{minipage}[t]{0.23\textwidth}
\caption{\KR{Inference speed- up achieved on different devices on Kimore 5 Test-set}}
\begin{tabular}{c c c} 
 \toprule
\multirow{2}{*}{Model} & GPU & \KR{CPU}\\
 & Time (s) & \KR{Time (s)}\\
\hline
STGCN\cite{deb2022graph} &  13.52 & \KR{85.68}\\ 
D-STGCNT\cite{mourchid2023d} & 2.05 & \KR{23.47} \\ 
\texttt{Ours} & 1.09 & \KR{10.08}\\
 \bottomrule

\end{tabular}
    \label{tab:IFS}
\end{minipage}
\end{table*}

\begin{figure*}[ht]
  \centering  
  \includegraphics[width=0.6\textheight]{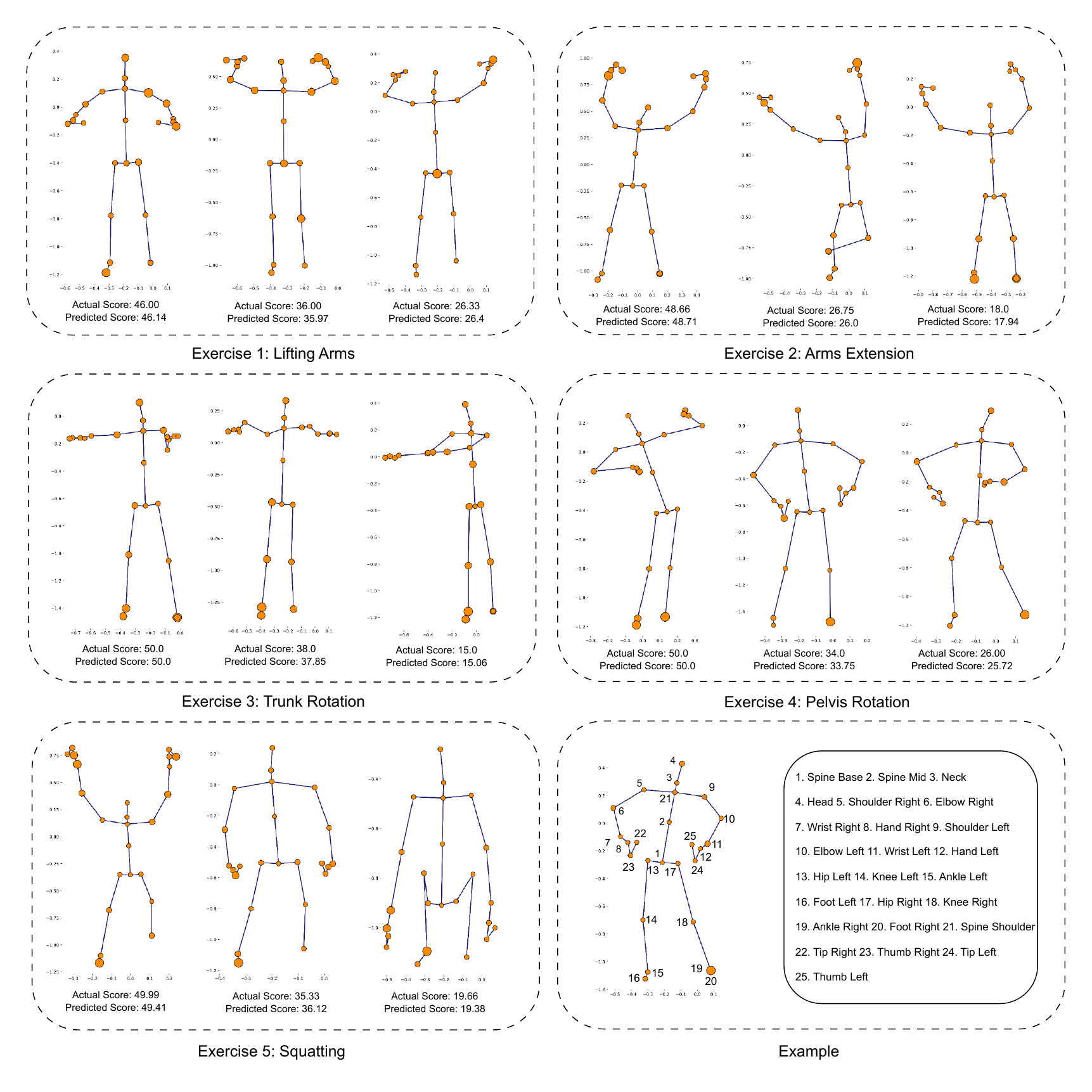}
  
  \caption{An illustration of positive integrated gradients calculated on each user joint for each exercise. The gradients are shown as joint size on the 2D skeleton diagrams. A larger joint showcases a higher attribution. We showcase different levels of accuracy for each type of exercise. \KR{We define expert, moderate, and inaccurate.} We label each joint on an example diagram. A higher joint attribution suggest the joint influences out regardless if the exercise is accurate or not. \KR{We also showcase the model does not overfit, look into non-joint patterns, or look into different joints for the same exercise when it is in a certain range.}}
\label{fig:IG}
\end{figure*}


\subsection{Experimental Results}
\label{sec:Experimental_Results}

In Table~\ref{tab:UI-PRMD}, Table~\ref{tab:IRDS}, and Table~\ref{tab:Kimore}, we report the experimental results of our system compared to other systems for the UI-PRMD, IRDS, and KIMORE datasets, respectively, for rehabilitation exercise assessment. For the KIMORE dataset, we report 3 separate performance metrics: MAD, RMSE, and MAPE. For the UI-PRMD dataset, we only measure performance using the MAD performance metric. We report the accuracy of the IRDS dataset. Performance was obtained by evaluating all the different model architectures in the test dataset. We compare our performance against different works to showcase the efficacy of our system. The experimental results for our study exceed the current SOTA results in every experiment and performance metric. The curve-based aggregation technique augments the point information, unlike STGCNs~\cite{deb2022graph} and GCNs\cite{song2020richly}, which cannot adequately incorporate the spatial and temporal knowledge of the point clouds. In addition to curve-based aggregation, attention modules can capture long-range and interpoint cloud information, surpassing models that introduce modules at the joint and frame level to capture better joint information and learn joint relationships~\cite{zhang2020semantics}. Neural networks specifically designed to capture spatiotemporal information~\cite{liao2020deep} do not perform well because the data points usually on the sample plane are topologically similar, so aggregation does not provide much spatial information.

\subsection{Robustness and Generalization Analysis}
\label{sec:Robustness Analysis}
 \KR{To validate the robustness of the proposed technique we conducted a Leave-One-Subject-Out (LOSO) cross-validation on the KIMORE dataset using the previously identified optimal hyperparameters. Here we carry out 5 experiments using different test-folds where each fold consists one expert, moderate and inaccurate instance. These experiments showcases the models generalizability and its adaptibility to redundancy and noise. By utilizing the optimal hyperparameter configuration identified during the initial tuning phase, this protocol ensures that the assessment performance is strictly subject-independent and free from the biases of a specific train-test partition. We show the result in Table \ref{tab:dev_locv}, yield a low standard deviation on every exercise, not higher than $0.07$. The low standard deviation across all subjects empirically confirms the architecture’s anatomical invariance, mainly the point-aggregation strategies ability to reduce redundencies and axial attention to relate to significant points, demonstrating its ability to provide consistent and accurate rehabilitation scores regardless of a participant's individual physical proportions or movement idiosyncrasies. This high level of stability underscores the system's readiness for practical, home-based deployment, where maintaining performance across a heterogeneous population without per-user calibration is essential. }\KRA{We further showcase, in Table \ref{tab:dev_all}, that our normal training on held-out user test sets also has low standard deviation in all exercises of every dataset across 10 experiments with a distinct random seed for each one.}

\begin{table}[htbp]
\setlength{\tabcolsep}{0pt} 
\caption{\KR{Mean and Standard Deviation across the KIMORE dataset exercises \KRA{on LOSO CV.}}}
\label{tab:dev_locv}
\begin{tabular*}{\columnwidth}{@{\extracolsep{\fill}} l c c c c c } 
\toprule
MAD & Ex 1 & Ex 2 & Ex 3 & Ex 4 & Ex 5 \\
\midrule
Mean  & 0.173 & 0.554 & 0.105 & 0.242 & 0.371 \\ 
Stdev & $\pm$ 0.04 & $\pm$ 0.03 & $\pm$ 0.07 & $\pm$ 0.05 & $\pm$ 0.05 \\
\bottomrule
\end{tabular*}

\label{tab:dev}
\end{table}

\begin{table*}[t]
\centering
\caption{\KRA{Per-exercise mean $\pm$ standard deviation of our method across datasets. KIMORE and UI-PRMD use MAD (lower is better); IRDS uses accuracy (higher is better).}}
\label{tab:dev_all}
\setlength{\tabcolsep}{4pt}
\resizebox{\textwidth}{!}{%
\begin{tabular}{l |c| c| c| c |c |c |c |c |c |c}
\toprule
Dataset (Metric) & Ex1 & Ex2 & Ex3 & Ex4 & Ex5 & Ex6 & Ex7 & Ex8 & Ex9 & Ex10 \\
\midrule
KIMORE (MAD) & 0.185\,$\pm$0.03 & 0.560\,$\pm$0.03 & 0.128\,$\pm$0.02 & 0.256\,$\pm$0.03 & 0.388\,$\pm$0.04 & - & - & - & - & - \\
UI-PRMD (MAD) & 0.008\,$\pm$0.0003 & 0.005\,$\pm$0.0006 & 0.006\,$\pm$0.0007 & 0.006\,$\pm$0.0007 & 0.006\,$\pm$0.0004 & 0.006\,$\pm$0.0002 & 0.010\,$\pm$0.0003 & 0.014\,$\pm$0.0004 & 0.008\,$\pm$0.0005 & 0.018\,$\pm$0.0001 \\
IRDS (Acc) & 1.000\,$\pm$0.01 & 0.957\,$\pm$0.04 & 0.979\,$\pm$0.01 & 0.974\,$\pm$0.01 & 0.986\,$\pm$0.03 & 0.985\,$\pm$0.01 & 1.000\,$\pm$0.02 & 0.971\,$\pm$0.02 & 0.986\,$\pm$0.01 & - \\
\bottomrule
\end{tabular}%
}
\end{table*}

\subsection{Ablation studies}    
\label{sec:Ablation_studies}

\noindent\textbf{Ablation Studies on Hyperparameters}: We vary the learning rate of the training pipeline from 0.01 to 0.00001 to adjust the learning rate so that the model learns the best. We find that the model performs best at a learning rate of 0.0001. 

\noindent\textbf{Internal Representation}: To explore internal representation learned by the model, we extract embeddings from the last transformer layer of the model and generate 2D t-distributed Stochastic Neighbor Embedding (t-SNE) embeddings for visualization on the test set of Kimore Ex 5. We seperate the samples in the test set into three categories: Accurate (scores greater than or equal to 40), Moderate (scores greater than or equal to 30 or less than 40), and inaccurate (scores less than 30). This is shown in Figure~\ref{fig:TSNE}. The t-SNE plot showcases a clear distinction between accurate and inaccurate samples. Clusters showcase model learns overall similar information to deduce accurate exercises. In contrast, the inaccurate exercise form a different cluster, indicating variability in the pattern learned for inaccurate samples. Moderate and accurate samples seemingly form a cluster. However, the samples in the cluster get more accurate moving leftwards on the graph. 

\noindent\textbf{Model Component Analysis}: We conduct extensive experiments on the nonutilization of different model components to understand each model component's efficacy in assessing the rehabilitation exercise. We also observe which model component impacts the overall performance of the model the most. Table~\ref{tab:UI-PRMD-Abl} shows the performance of the model with different model components removed for UI-PRMD. We experiment and evaluate the model without the curve-based aggregation module and transformer layers. Performance improves drastically by adding the CurveNet block, as it increases data points according to its position. We also carry out experiments without the first and second transformer layers. We notice a drop in the model's performance on the UI-PRMD dataset without any transformer layers missing. More transformer layers tend to capture all the critical spatial and temporal features that allow the model to predict better. 

\noindent\textbf{Inference}: \KR{We compare our model's inference speed against popular methods in Table~\ref{tab:IFS}. We report an inference time of 1.09 seconds on 75 instances (totaling 7,500 frames). 
Time per Instance: $1.09s / 75 \approx \mathbf{15 ms}$ per exercise video.
Time per Frame: $1.09s / 7500 \approx \mathbf{0.15 ms}$ per frame.
While traditional methods like STGCN rely on heavy graph adjacency matrix multiplications that scale poorly, our CurveNet+Transformer approach simplifies the skeleton data into geometric curves, significantly reducing the total floating-point operations (FLOPs).}

\KR{On an Nvidia RTX 3070 Ti GPU, our model processes the entire test set in 1.09s, achieving a 12.4x speedup over the baseline. Furthermore, we evaluate the potential for edge deployment by analyzing CPU inference. Unlike GCNs, which often suffer from irregular memory access on CPUs, the Transformer’s axial-attention mechanism is highly optimized for modern CPU instruction sets. We showcase in Table~\ref{tab:IFS} a CPU inference time of 10.08s on a consumer-grade CPU (Core i7-12700 12th Gen). Even without GPU acceleration, our model maintains a high throughput, making it suitable for real-time rehabilitation monitoring in home-based settings where specialized hardware may be unavailable.}

\subsection{Remote Monitoring Evaluation}

\KRA{The proposed framework monitors a patient's movement utilizing a model trained on expert-annotated data. This allows users to carry out rehabilitation exercises without expert supervision remotely. The system in this study deals with precomputed kinematic data through Kinect and Vicon sensors. For this evaluation, we assume precomputed, quasi-systems surrounding our model. We also assume jitter, packet-loss, angular, trajectory accuracy, and similar measurements in accordance with Kinect and Vicon sensors which is relative. We also characterise the system along three axes: the rate at which the system ingests data, the speed at which it produces a score, and the communication cost of transmitting the skeleton stream from the patient to the clinic (including surrounding systems and our model in federation). We discuss the actions on the Kimore dataset.}

\KRA{\noindent\textbf{Acquisition and processing:} The input is acquired with a Microsoft Kinect sensor at a frame rate of $30$~fps. For KIMORE, each frame contains $25$ joints, giving a sampling rate of $25\times30=750$ joint measurements per second. Because a home setting cannot assume a dedicated GPU, we focus on CPU performance. On a consumer CPU (Core~i7-12700), the model processes the entire test set ($75$ videos, $7{,}500$ frames) in $10.08$~s, corresponding to approximately $1.34$~ms per frame and $134$~ms per exercise video. As this per-frame time is far below the $33.3$~ms inter-frame interval at $30$~fps, the system sustains real-time throughput on the CPU alone, with roughly $25\times$ headroom and no specialised hardware. For reference, a GPU (RTX~3070~Ti) further reduces this to $\sim$0.15~ms per frame ($\sim$15~ms per video), but is not required for real-time operation.}

\KRA{\noindent\textbf{Communication cost:} The system transmits only skeletal coordinates rather than RGB or depth video, which makes the stream extremely lightweight. Each frame carries $25\ \text{joints}\times3\ \text{coordinates}\times4\ \text{bytes}=300$~bytes, so the sustained data rate is $300\times30\approx9$~KB/s ($\approx72$~kbps). This is well within the uplink capacity of most home broadband connection. On a modest $10$~Mbps uplink, the transmission time of one second of data is about $7$~ms; adding a typical wide-area round-trip time ($30$--$50$~ms) and the $\sim$15~ms inference latency yields an estimated end-to-end latency of roughly $50$--$70$~ms, below the $\sim$100~ms threshold commonly associated with perceived real-time feedback.}

\begin{table}[t]
\centering
\caption{\KRA{System and communication characteristics for remote monitoring (KIMORE, skeleton stream).}}
\label{tab:remote}
\begin{tabular}{lc}
\toprule
Parameter & Value \\
\midrule
Frame rate                       & 30 fps \\
Sampling rate                    & 750 joint measurements/s \\
Processing time per frame        & $\sim$0.15 ms \\
Processing time per exercise     & $\sim$15 ms \\
Payload per frame                & 300 bytes \\
Required throughput              & $\sim$72 kbps \\
Transmission time (10 Mbps)      & $\sim$7 ms \\
Estimated end-to-end latency     & $\sim$50--70 ms \\
\bottomrule
\end{tabular}
\end{table}

\KRA{\noindent\textbf{Scope.} The proposed model consumes sensor-provided skeleton data and outputs a quality score; motion-measurement accuracy (e.g.\ velocity or trajectory estimation) is therefore a property of the capture device rather than of the assessment model. The communication figures above are analytical estimates based on the skeleton payload size and typical home-network conditions. Sustained jitter, packet loss, and data availability depend on the deployment network and require an end-to-end live deployment, which we identify as future work. We further add our system leaves room for packet loss coverage and augmentation to minimize jitters.}

\KR{\noindent\textbf{Feature-Based Explanation}: In addition to performance evaluation, we aim to provide an interpretable explanation of which body joints most influence the predicted exercise quality score. To this end, we employ positive Integrated Gradients (IG) as a feature-attribution method. Unlike signed attribution methods that distinguish between positive and negative contributions, our approach focuses solely on attribution magnitude, indicating how strongly each joint affects the model’s prediction, rather than in which direction.
Figure~\ref{fig:IG} visualizes the joint-level attributions for three representative accuracy levels: expert, moderate, and inaccurate; for each exercise. Joint importance is encoded using both color intensity and joint size, where darker color and larger diameter indicate higher attribution. These values represent the relative influence of each joint on the final score, not whether the joint improves or degrades performance.
Importantly, a high positive attribution does not imply correctness. Instead, it indicates that the joint is critical to the model’s decision, regardless of whether the predicted score is high or low. For example, a joint with high attribution in an inaccurate execution suggests that its configuration is a major factor contributing to the low score. Conversely, high attribution in an expert execution indicates that the joint configuration is key to achieving a high-quality score.
Across different exercises, we observe that distinct sets of joints consistently receive high attribution, demonstrating that the model has learned exercise-specific motion patterns rather than environment-dependent cues. For the lifting arms exercise (Exercise 1), the left shoulder, left elbow, left wrist, and right elbow exhibit high attribution for both expert and moderate executions, while deviations in spine and wrist positions dominate the attribution in inaccurate executions. Similarly, for trunk rotation (Exercise 3), the head, shoulders, hips, and ankles are consistently influential across all accuracy levels, with spine alignment playing a critical role in inaccurate executions.
In some cases, the spine receives lower attribution despite being biomechanically important. This may be due to gradient saturation, feature interactions, or the fact that the spine configuration is less informative at certain phases of the exercise. Overall, these results indicate that the proposed explainability mechanism highlights where the model focuses its attention, thereby offering actionable insight into which joints users should monitor or correct during exercise execution.
The method provides actionable insight by identifying which joints are responsible for the evaluation outcome. In accurate executions, high attribution highlights joints that are correctly engaged, whereas in inaccurate executions, high attribution localizes the source of error. For example, in the raising hands exercise, arm joints dominate attribution in expert executions, while increased attribution of the spine and wrists in inaccurate executions indicates postural compensation or misalignment. This allows users to focus their correction efforts on specific body parts and refer back to experts or expert videos or self-improvement towards only specific joints which makes understanding and debugging exercise postures easy.}

\noindent\textbf{Limitations:}
\KRA{While our model achieves state-of-the-art results, several limitations remain. First, we did not perform cross-dataset validation because the three datasets use different sensors, joint counts, exercise sets, and scoring conventions, so a naive transfer would conflate domain shift with genuine generalization failure; a shared joint mapping and score normalization is left as future work. Second, the reported communication figures are analytical estimates, and a live clinical deployment is required to measure real jitter, packet loss, and availability. Third, limited compute prevented scaling to larger model variants, and we did not visualize cross-feature relations in the explainability analysis.}

\section{Conclusion}
\label{sec:Conclusion}

This work proposes a transformer network that utilizes joint coordinates as point clouds to assess physical rehabilitation exercise. We augment joint coordinates to integrate descriptive trainable features to enhance the spatio-temporal information of the point cloud. We further utilize a transformer network with axial attention layers that calculate spatio-temporal attention between coordinates and between timesteps. Models are trained to accurately rate the skeleton coordinates to perform physical rehabilitation exercises properly. We also visualize different body joints and their roles so that users can learn which joints they have to put more focus on or improve. The efficacy of the model comes from the utilization and availability of all the model components. Our model achieves state-of-the-art performance and outperforms existing models. \KR{For future work we look understand a full biomechanical analysis of cross-feature relations and kinetic synergies of the model. We look to further our research by developing a helpful UI system based on the features and their synergies to help patients better understand explanations surrounding wrong exercises.}

\vspace{1em}
\noindent\textbf{Acknowledgment:} Authors acknowledge the support from Apurba-NSU R\&D Lab, Department of Electrical and Computer Engineering, North South University, Dhaka, Bangladesh, and Conference Travel and Research Grants (CTRG) 2023-2024 from North South University, under Grant ID: CTRG-23-SEPS-25.

\bibliographystyle{ieeetr} 
\bibliography{biblio.bib}

\end{document}